\documentclass[manuscripts，review]{acmart}
\usepackage{color}
\usepackage{multirow}
\usepackage{xspace}
\usepackage{float}
\usepackage[ruled,vlined]{algorithm2e}
\definecolor{Gray}{gray}{0.95}
\newcommand{\ours}{Agentic Lybic\xspace}
\AtBeginDocument{%
  }

\copyrightyear{2025}
\acmYear{2026}
\setcopyright{cc}
\setcctype{by}
\acmConference[CHI '26]{CHI Conference on Human Factors in Computing Systems}{April 13–17, 2026}{Creant el demà junts, Barcelona}
\acmBooktitle{CHI Conference on Human Factors in Computing Systems (CHI '26), April 13–17, 2026, Creant el demà junts, Barcelona}
\acmISBN{978-1-4503-XXXX-X/2026/04}

\begin{document}

\title{\ours: Multi-Agent Execution System with Tiered Reasoning and Orchestration}

\author{Liangxuan Guo}
\authornotemark[1]
\email{guoliangxuan@deepmatrix.com.cn}
\affiliation{%
  \institution{Lybic}
  \city{Beijing}
  \country{China}
}

\author{Bin Zhu}
\authornotemark[1]
\email{zhubin@tingyutech.com}
\affiliation{%
  \institution{Lybic}
  \city{Beijing}
  \country{China}
}

\author{Qingqian Tao}
\authornotemark[1]
\email{taoqingqian@tingyutech.com}
\affiliation{%
  \institution{Lybic}
  \city{Beijing}
  \country{China}
}

\author{Kangning Liu}
\authornotemark[1]
\email{liukangning@deepmatrix.com.cn}
\affiliation{%
  \institution{Lybic}
  \city{Beijing}
  \country{China}
}

\author{Xun Zhao}
\authornotemark[1]
\email{zhaoxun@deepmatrix.com.cn}
\affiliation{%
  \institution{Lybic}
  \city{Beijing}
  \country{China}
}

\author{Xianzhe Qin}
\authornotemark[1]
\email{qinxianzhe@deepmatrix.com.cn}
\affiliation{%
  \institution{Lybic}
  \city{Beijing}
  \country{China}
}

\author{Jin Gao}
\authornotemark[1]
\authornotemark[1]
\email{gaojin@deepmatrix.com.cn}
\affiliation{%
  \institution{Lybic}
  \city{Beijing}
  \country{China}
}

\author{Guangfu Hao}
\authornotemark[1]
\email{haoguangfu@deepmatrix.com.cn}
\affiliation{%
  \institution{Lybic}
  \city{Beijing}
  \country{China}
}

\authornote{These authors contributed equally to this work. **Corresponding author(s): Guangfu Hao (haoguangfu@deepmatrix.com.cn), Jin Gao (gaojin@deepmatrix.com.cn)}

\renewcommand{\shortauthors}{Guo et al.}

\begin{abstract}
  Autonomous agents for desktop automation struggle with complex multi-step tasks due to poor coordination and inadequate quality control. We introduce \textsc{Agentic Lybic}, a novel multi-agent system where the entire architecture operates as a finite-state machine (FSM). This core innovation enables dynamic orchestration. Our system comprises four components: a Controller, a Manager, three Workers (Technician for code-based operations, Operator for GUI interactions, and Analyst for decision support), and an Evaluator. The critical mechanism is the FSM-based routing between these components, which provides flexibility and generalization by dynamically selecting the optimal execution strategy for each subtask. This principled orchestration, combined with robust quality gating, enables adaptive replanning and error recovery. Evaluated officially on the OSWorld benchmark, \textsc{Agentic Lybic} achieves a state-of-the-art 57.07\% success rate in 50 steps, substantially outperforming existing methods. Results demonstrate that principled multi-agent orchestration with continuous quality control provides superior reliability for generalized desktop automation in complex computing environments.

\end{abstract}


\begin{teaserfigure}
  \includegraphics[width=\textwidth]{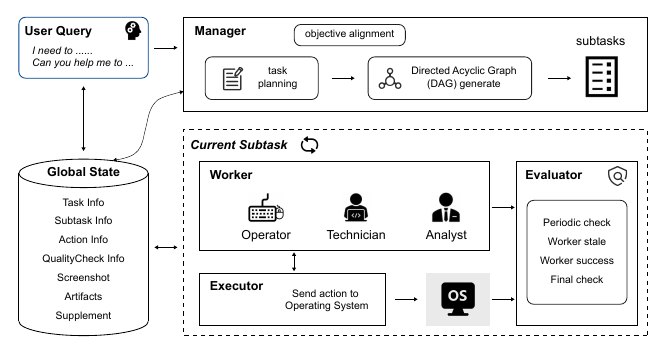}
  \caption{Overview of the Agentic Lybic multi-agent system architecture.}
  \Description{The Manager decomposes user queries into subtasks using DAG planning. Specialized Workers (Operator, Technician, Analyst) execute actions through the Executor, while the Evaluator provides continuous quality monitoring. The Global State maintains system information and enables dynamic coordination between components.}
  \label{fig:teaser}
\end{teaserfigure}


\maketitle

\section{Introduction}

The automation of complex desktop tasks through autonomous agents represents one of the most challenging frontiers in artificial intelligence, requiring systems that can navigate intricate multi-step workflows while maintaining reliability and efficiency\cite{zhang2025largelanguagemodelbrainedgui, nguyen2025guiagentssurvey}. Recent advances in computer-using agents have demonstrated promising capabilities in executing tasks through Graphical User Interfaces (GUIs), with vision-language models enabling increasingly sophisticated interactions with visual elements\cite{gemini25,openAI_o3_o4_mini,openai2024gpt4ocard, cua2025, qin2025ui}. Current approaches to desktop automation typically fall into two categories: GUI-centric agents that rely exclusively on visual interaction\cite{agashe2025agents2compositionalgeneralistspecialist,yang2025gta1guitesttimescaling}, and hybrid systems that combine GUI manipulation with programmatic execution. While GUI-only agents offer intuitive human-like interaction patterns, they suffer from brittleness in complex scenarios due to visual grounding ambiguity and accumulated error propagation over long sequences. Recent hybrid approaches, such as CoAct-1\cite{song2025coact1computerusingagentscoding}, have addressed some of these limitations by introducing specialized programming agents alongside GUI operators, achieving notable improvements in both efficiency and success rates. However, these systems' core limitation lies in their handling of long-horizon tasks. They often employ a simplistic "delegate-and-forget" approach, which lacks the continuous oversight and adaptive re-planning needed for robust error recovery. Consequently, they fail to implement the sophisticated routing mechanisms and comprehensive quality control essential for coordinating multiple functional modules over extended workflows.

The core challenge lies not merely in expanding the action space of computer-using agents, but in orchestrating multiple specialized components through principled coordination mechanisms that can dynamically adapt to changing task requirements and handle execution failures gracefully\cite{tallam2025autonomous}. Existing systems often treat task decomposition and execution as separate phases, lacking the continuous feedback loops necessary for robust long-horizon performance\cite{hu2024hiagent}. Furthermore, current quality assessment approaches are typically binary and reactive, missing opportunities for proactive intervention\cite{sun2025assistantx} and incremental course correction.

To address these issues above, we draw inspiration from finite-state machines (FSMs), which provide a structured and predictable framework for managing complex, state-dependent workflows. In this work, we introduce \textbf{Agentic Lybic}, a novel FSM-based multi-agent execution system, which could addresse these fundamental limitations through a tiered reasoning architecture and sophisticated orchestration framework. Our system advances beyond simple hybrid execution by implementing a comprehensive four-tier architecture: a \textbf{Controller} that manages global state and decision triggers, a \textbf{Manager} for intelligent task decomposition and adaptive re-planning, a \textbf{Worker} subsystem with three specialized roles (Technician for system operations, Operator for GUI interactions, and Analyst for decision support), and an \textbf{Evaluator} that provides continuous quality assessment and intervention triggers.

The key innovation of our approach lies in the dynamic orchestration mechanism (i.e., routing) that seamlessly coordinates between different functional modules based on ongoing task assessment, coupled with a comprehensive quality gate system that enables persistent monitoring, adaptive re-planning, and robust error recovery. Unlike existing systems that perform static task delegation, Agentic Lybic implements a feedback-driven execution model where performance is continuously monitored through multiple trigger mechanisms (periodic checks, stagnation detection, and success verification), enabling the system to adaptively adjust its strategy mid-execution.

Our tiered reasoning approach introduces several novel contributions: (1) a state-aware orchestration framework that dynamically selects optimal execution strategies based on task characteristics and current system state, (2) a comprehensive quality gate system with multiple intervention triggers that enable proactive error handling and adaptive re-planning, (3) a specialized worker architecture that provides fine-grained control over different execution modalities while maintaining seamless coordination, and (4) an incremental clarification policy that systematically addresses visual ambiguity in GUI-dense environments.

We evaluate Agentic Lybic on the challenging OSWorld benchmark, where our system achieves a new state-of-the-art success rate of 57.07\% in 50 steps, representing a substantial improvement over existing methods including the recent CoAct-1 \citep{song2025coact1computerusingagentscoding} (56.4\%) and agent s2.5 \citep{agashe2025agents2compositionalgeneralistspecialist} (54.2\%). Beyond raw performance gains, our system demonstrates superior reliability in long-horizon scenarios and maintains efficiency through proactive quality control. Our results demonstrate that principled multi-agent orchestration with continuous quality assessment provides a more robust and scalable foundation for generalized desktop automation, opening new possibilities for autonomous task execution in complex computing environments. The tiered reasoning approach not only improves success rates but also provides a systematic framework for handling the inherent complexity and uncertainty of real-world desktop automation tasks.

\section{Related Work}

\subsection{Screen Understanding and Visual Grounding}

A foundational challenge in GUI automation lies in accurately perceiving and grounding interface elements from raw pixel input. Early approaches focused on developing sophisticated screen parsing capabilities that could identify and locate interactive elements without relying on structured representations like DOM trees or accessibility hooks\cite{lai2024autoweb}. OmniParser~\citep{lu2024omniparserpurevisionbased,yu2025omniparser} pioneered this direction by learning screen-parsing primitives for pure vision-based understanding, enhancing the ability of large models to generate accurate actions in interface regions through interactive icon detection and semantic element extraction.

The grounding problem—mapping natural language instructions to actionable screen locations—has been addressed through several specialized systems. SeeClick introduced instruction-to-target grounding capabilities~\citep{cheng2024seeclick}, while Aria-UI~\citep{yang2024aria} and UGround~\citep{gou2024navigating} extended this to universal GUI grounding across diverse interfaces. OS-Atlas~\citep{wu2024atlas} represents a major advance in this area, training a foundation action model that generalizes across multiple platforms (Windows, Linux, MacOS, Android, and web) using a massive corpus of over 13 million GUI elements. The system demonstrates how large-scale, cross-platform training data can substantially improve GUI grounding performance, particularly in out-of-distribution scenarios. ScreenSpot-Pro~\citep{li2025screenspot} provides dedicated grounding evaluation benchmarks under professional, high-resolution settings, addressing the growing need for robust performance assessment in complex visual environments.

\subsection{End-to-End GUI Agents}

The end-to-end paradigm represents a paradigmatic shift toward unified models that integrate perception, reasoning, and action generation within a single model. These approaches aim to eliminate the need for separate planning and grounding components by training models that can directly predict executable actions from visual input and high-level instructions.

CogAgent~\citep{hong2024cogagent} exemplifies this approach as an 18B parameter visual language model specifically designed for GUI understanding and navigation. By utilizing both low-resolution and high-resolution image encoders, CogAgent supports input at 1120×1120 resolution, enabling recognition of tiny page elements and text. The model achieves state-of-the-art performance on multiple text-rich and general VQA benchmarks while outperforming LLM-based methods on both PC and Android GUI navigation tasks using only screenshots as input. More recently, GUI-Owl~\citep{ye2025mobileagentv3fundamentalagentsgui} has pushed the boundaries of end-to-end GUI agents through three key innovations: large-scale environment infrastructure enabling self-evolving trajectory production, diverse foundational agent capabilities integrating UI grounding with planning and reasoning, and scalable environment reinforcement learning for real-world alignment. GUI-Owl-7B achieves impressive performance scores of 66.4 on AndroidWorld and 29.4 on OSWorld, demonstrating the potential of purpose-built foundational models for GUI automation.

UI-TARS~\citep{qin2025ui} represents a native end-to-end GUI agent model that processes raw screenshots and generates human-like interactions through unified perception, reasoning, action, and memory capabilities. The system integrates enhanced visual understanding through large-scale GUI datasets, unified action modeling across platforms, System-2 reasoning with explicit thought generation, and iterative refinement via online trace collection. UI-TARS-2~\citep{wang2025uitars2technicalreportadvancing} further advances this approach through multi-turn reinforcement learning and hybrid GUI environments that combine screen actions with file system access. AGUVIS~\citep{xu2024aguvis} and InfiGUIAgent~\citep{liu2025infiguiagent} enhance GUI agent autonomy through unified visual frameworks, with InfiGUIAgent introducing a two-stage training pipeline that advances GUI task automation using inner monologue techniques. GUI-R1~\citep{luo2025gui} demonstrates how rule-based reinforcement fine-tuning can enhance high-level GUI action prediction while achieving competitive performance with considerably less training data. InfiGUI-R1~\citep{liu2025infigui} shifts agents from reactive acting to deliberative reasoning through reasoning spatial distillation and reinforcement learning approaches.

Recent work has also focused on addressing temporal dynamics and action prediction capabilities. ScaleTrack~\citep{huang2025scaletrack} predicts future actions from current GUI images and backtracks historical actions, thereby explaining the evolving correspondence between GUI elements and actions. UITron-Speech~\citep{zeng2025uitronfoundationalguiagent} extends beyond text-based instructions by introducing the first end-to-end GUI agent capable of processing speech instructions directly, utilizing mixed-modality training strategies and two-step grounding refinement to handle the inherent challenges of speech-driven interface interaction.

\subsection{Multi-Agent Frameworks}

While end-to-end models show promise for unified GUI interaction, agentic frameworks focus on orchestrating multiple specialized components to leverage complementary strengths and achieve more robust performance on complex tasks. These systems typically separate high-level planning from low-level execution while introducing sophisticated coordination mechanisms. The modular planner-grounder paradigm explicitly separates "what to do" from "where and how to act on screen." Representative systems like SeeClick~\citep{cheng2024seeclick} and OS-Atlas~\citep{wu2024osatlasfoundationactionmodel} demonstrate how language planners can propose subgoals while visual models handle grounding. GTA-1~\citep{yang2025gta1guitesttimescaling} strengthens this approach through test-time scaling, sampling multiple candidate actions and using multimodal large language model~(MLLM) judges for selection, improving robustness on high-resolution, cluttered interfaces.

CoAct-1~\citep{song2025coact1computerusingagentscoding} represents the most recent advance in hybrid agentic frameworks, introducing a paradigm where agents can use coding as an enhanced action modality. The system features an Orchestrator that dynamically delegates subtasks between a GUI Operator and a specialized Programmer agent capable of writing and executing Python or Bash scripts. This hybrid approach demonstrates substantial performance improvements by leveraging the complementary strengths of GUI manipulation and programmatic execution, achieving state-of-the-art results on the OSWorld benchmark.

Beyond GUI-specific systems, several frameworks provide general infrastructure for multi-agent orchestration and tool composition. Agent-S/Agent-S2~\citep{agashe2024agent,agashe2025agent} and AutoGen~\citep{wu2024autogen} offer reusable infrastructures for multi-agent coordination and tool calling. UFO-2~\citep{zhang2025ufo2}, PyVision~\citep{zhao2025pyvision}, and ALITA~\citep{qiu2025alita} extend this principle to dynamic tool construction and invocation, though they are not specifically designed for GUI automation.

While these agentic frameworks have made significant progress in combining different execution modalities, they often lack sophisticated quality control mechanisms and continuous oversight capabilities. Most systems treat task decomposition and execution as largely separate phases, missing opportunities for adaptive re-planning and proactive error correction that are essential for robust long-horizon performance. Our work addresses these limitations through a self-consistent FSM-based tiered reasoning architecture with continuous quality assessment and dynamic orchestration capabilities.

\section{Agentic Lybic: FSM-based Multi-Agent Architecture}

In this section, we present the detailed design of \textsc{Agentic Lybic}. Our system transcends traditional approaches by implementing a four-tier architecture with continuous quality control and adaptive re-planning mechanisms, which is the core of state transition of the FSM.

\subsection{System Architecture Overview}

\begin{figure}[!t]
  \includegraphics[width=\textwidth]{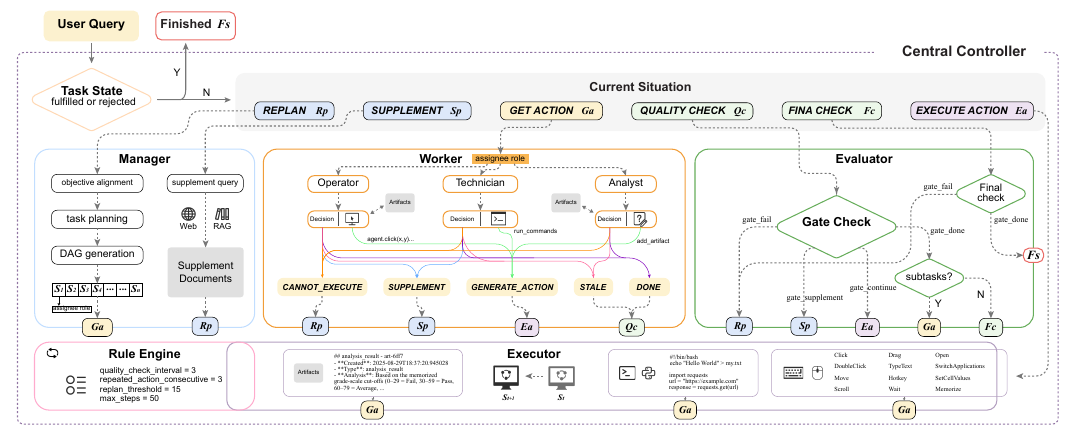}
  \caption{State transition diagram of \textsc{Agentic Lybic} showing the tiered orchestration workflow. The Central Controller manages six core situations (REPLAN, SUPPLEMENT, GET ACTION, QUALITY CHECK, FINAL CHECK, EXECUTE ACTION) with dynamic transitions based on execution outcomes. The Manager handles task decomposition and re-planning, the Worker subsystem provides specialized execution through three roles (Operator for GUI, Technician for system operations, Analyst for decision support), and the Evaluator implements comprehensive quality gates with multiple trigger mechanisms (periodic checks, stagnation detection, success verification).}
  \label{fig:architecture_flow}
\end{figure}

\textsc{Agentic Lybic} is built upon a hierarchical four-component architecture designed to maximize coordination efficiency while maintaining robust execution control (Figure~\ref{fig:architecture_flow}). The system comprises: (1) a \textbf{Controller} that manages global state transitions and decision triggers, (2) a \textbf{Manager} responsible for intelligent task decomposition and adaptive re-planning, (3) a \textbf{Worker} subsystem with three specialized execution roles, and (4) an \textbf{Evaluator} that provides continuous quality assessment and intervention mechanisms.

The key innovation lies in our state-driven orchestration framework (i.e., state transition), where each component operates within a well-defined state space that enables seamless coordination and robust error handling. Unlike existing systems that rely on simple delegation patterns, our architecture implements a continuous feedback loop with multiple quality gates that enable proactive intervention and adaptive strategy adjustment.

\subsection{Central Controller: State Management and Decision Orchestration}

The Controller serves as the central nervous system of our architecture, managing global state transitions and orchestrating component interactions through a sophisticated state machine. We define six primary controller situations that capture the essential phases of task execution:

\begin{itemize}
\item \textbf{REPLAN ($R_p$)}: Triggered when task initial decomposition or strategy adjustment is required
\item \textbf{SUPPLEMENT ($S_p$)}: Activated when additional contextual information from Web or external knowledge base is needed
\item \textbf{GET\_ACTION ($G_a$)}: Gets specific action generation by Worker components
\item \textbf{QUALITY\_CHECK ($Q_c$)}: Evaluates execution effectiveness and determines continuation strategy
\item \textbf{FINAL\_CHECK ($F_c$)}: Performs comprehensive task completion verification
\item \textbf{EXECUTE\_ACTION ($E_a$)}: Coordinates actual operation execution through the hardware interface
\end{itemize}

The Controller maintains a comprehensive state space that encompasses task-level status (\texttt{created, pending, on\_hold, fulfilled, rejected}), subtask-level progress (\texttt{ready, pending, stale, fulfilled, rejected}), and execution-level outcomes (\texttt{executed, timeout, blocked, error}). This multi-granular state representation enables fine-grained control over the execution process while providing robust error handling capabilities.

Formally, we define the global state as a tuple $S = (S_T, S_{ST}, S_E, C)$, where $S_T$ represents task status, $S_{ST}$ denotes subtask status, $S_E$ captures execution status, and $C$ indicates the current controller situation. The state transition function is defined as:

\begin{equation}
S_{t+1} = \delta(S_t, A_t, O_t)
\end{equation}

where $A_t$ represents the action taken at time $t$, $O_t$ is the observation received, and $\delta$ is the state transition function (i.e., the whole \textsc{Agentic Lybic} system) that determines the next state based on current context and execution outcomes.

The Controller employs a trigger code system (i.e., a state transition look-up-table) organized into ten primary categories that enable precise coordination between components and robust error handling (see Table~\ref{tab:trigger_codes} for complete reference). This trigger-driven architecture encompasses: Rule Validation triggers for automated quality control (periodic checks, stagnation detection, execution limits), Task Status Rules for system-wide oversight and resource management (completion detection, runtime limits, state switch boundaries), Worker Coordination triggers for managing the interface between planning and execution across all three Worker roles, and Error Recovery mechanisms for graceful degradation under unexpected conditions. The comprehensive trigger system ensures that Agentic Lybic maintains precise control over execution flow while providing the flexibility necessary for handling diverse desktop automation scenarios, with each trigger category contributing to different aspects of system reliability and coordination effectiveness.

\subsection{Manager: Task Decomposition and Adaptive Re-planning}

The Manager component implements advanced planning capabilities that go beyond simple task decomposition. Followed by Agent-S/Agent-S2~\citep{agashe2024agent,agashe2025agent}, it employs a directed acyclic graph (DAG) representation for subtask dependencies, enabling sophisticated scheduling and parallel execution opportunities. The Manager's core responsibilities include:

\textbf{Objective Alignment}: The Manager begins by analyzing the user's high-level intent and aligning it with the current visual context captured through screenshots. This alignment process disambiguates user queries by grounding abstract intentions in the concrete desktop environment, identifying available applications, current system state, and potential execution pathways.

\textbf{Initial Planning Generation}: Following objective alignment, the Manager generates a comprehensive initial plan containing all necessary subtasks to achieve the user's goal. This initial planning phase creates subtask specifications without dependency constraints, focusing on completeness and task coverage.

\textbf{DAG Construction}: The Manager then transforms the initial plan into a directed acyclic graph (DAG) representation with explicit structure:

\begin{itemize}

\item \textbf{NODES}: Each node contains a subtask title, detailed description, and assigned worker role (Technician for system operations, Operator for GUI interactions, or Analyst for decision support)

\item \textbf{EDGES}: Directed connections between nodes representing execution dependencies and precedence constraints

\end{itemize}

Based on the DAG structure, the Manager performs topological sorting to generate the actual execution sequence, ensuring subtask execution respects dependency constraints while identifying opportunities for parallel execution.

\textbf{Adaptive Planning}: The system implements dynamic planning adjustment through structured prompting strategies rather than mathematical optimization. The adjustment mechanism employs three levels: (1) \textit{light adjustment} for parameter modifications within existing subtasks, (2) \textit{medium adjustment} for subtask reordering and dependency restructuring, and (3) \textit{heavy adjustment} for complete task re-decomposition when fundamental strategy changes are required. These adjustments are triggered by execution feedback and implemented through carefully crafted prompt templates that guide the Manager toward appropriate planning modifications.

\textbf{Supplement Integration}: The Manager incorporates a supplement mechanism that leverages external knowledge sources (web search, Retrieval-Augmented Generation (RAG) systems) to enhance task understanding and fill information gaps. This capability is particularly crucial for handling novel scenarios or domain-specific requirements that may not be covered in the base knowledge.

\subsection{Worker Subsystem: Specialized Multi-Modal Execution}

The Worker subsystem represents a significant advancement over traditional single-modality approaches by implementing three specialized execution roles, each optimized for specific types of operations:

\textbf{Operator}: Manages GUI-based interactions using vision-language models for visual grounding and action generation. The Operator excels in scenarios requiring human-like interface navigation, form filling, visual content interpretation, and any tasks where GUI interaction is the primary or only available interface. It implements sophisticated visual grounding techniques to handle dense GUI environments and maintains context awareness across multi-step interaction sequences.

The Operator supports a comprehensive action repertoire including fundamental mouse operations (Click, DoubleClick, Move, Drag), keyboard interactions (TypeText, Hotkey), navigation controls (Scroll, SwitchApplications), and specialized functions for different contexts (SetCellValues for spreadsheets, Open for file operations). Additionally, it provides system coordination capabilities through Screenshot for visual state capture, Wait for timing control, and a unique Memorize function that enables cross-component information sharing by writing contextual memories to shared artifacts for other modules to access. Task completion is managed through explicit Done and Failed signals that trigger appropriate state transitions in the orchestration framework.

\textbf{Technician}: Handles system-level operations through terminal commands and script execution. The Technician is particularly effective for file system operations, environment configuration, batch processing, and any tasks that can be accomplished more reliably through programmatic interfaces than GUI manipulation. It supports both Python and Bash scripting environments and implements secure execution boundaries to prevent system compromise.

\textbf{Analyst}: Provides decision support and analytical capabilities for complex reasoning tasks. The Analyst is particularly essential for complex workflows where the Operator alone cannot complete the full task sequence—such as examination or assessment scenarios where questions must first be collected, analyzed, and answered before responses can be input. In these multi-stage workflows, the Operator first captures question content and writes it to shared artifacts, the Analyst retrieves this information from artifacts to perform reasoning and generate answers, then writes the solutions back to artifacts, enabling the Operator to subsequently extract and input the final answers. This artifact-mediated collaboration allows the system to handle sophisticated question-answering tasks that require separation of perception, reasoning, and action phases.

\subsection{Evaluator: Continuous Quality Assessment and Intervention}

The Evaluator component implements a comprehensive quality control framework that represents one of our key innovations. Unlike traditional binary success/failure assessments, our Evaluator provides continuous monitoring with multiple intervention triggers:

\textbf{Gate Decision Framework}: The Evaluator employs a comprehensive gate decision mechanism with four possible outcomes: \texttt{gate\_done} (subtask completed successfully), \texttt{gate\_fail} (execution failed, requires re-planning), \texttt{gate\_continue} (execution in progress, continue current strategy), and \texttt{gate\_supplement} (additional information needed).

\textbf{Multi-Trigger Quality Assessment}: The system implements three distinct trigger mechanisms for quality evaluation:
\begin{itemize}
\item \textit{PERIODIC\_CHECK}: Regular assessment every 5 execution steps to ensure consistent progress and stagnation detection when identical actions are repeated more than 3 times or single subtask execution exceeds 15 actions.
\item \textit{WORKER\_STALE}: Triggered when workers report they are unable to determine how to continue execution.
\item \textit{WORKER\_SUCCESS}: Verification trigger when workers report task completion.
\end{itemize}

The gate decision function processes these triggers through a comprehensive evaluation:

\begin{equation}
G(s, v_t, v_{t-1}) = \begin{cases}
\text{gate\_done} & \text{if } \text{similarity}(v_t, v_{target}) > \tau_{done} \\
\text{gate\_fail} & \text{if } \text{progress}(v_t, v_{t-1}) < \tau_{fail} \\
\text{gate\_continue} & \text{if } \tau_{fail} \leq \text{progress}(v_t, v_{t-1}) < \tau_{done} \\
\text{gate\_supplement} & \text{if } \text{uncertainty}(s) > \tau_{supplement}
\end{cases}
\end{equation}

where $v_t$ represents the current visual state, $v_{target}$ is the expected target state, and $s$ denotes the current system state. All these functions described above (e.g., $\text{similarity}(\cdot,\cdot)$) are all served by the Evaluator role.

\textbf{Final Check Verification}: The Evaluator implements a comprehensive final verification mechanism that activates when all subtasks reach completion status, serving as the ultimate quality gate before task termination. This final check performs holistic assessment beyond individual subtask verification, examining whether the combined execution results satisfy the user's original intent. The system supports five outcomes: final\_check\_passed (successful completion), final\_check\_failed (objectives unmet, triggers re-planning), final\_check\_pending (additional work discovered), final\_check\_error (verification error, triggers termination), and task\_impossible (clean termination for intractable tasks).

\subsection{Workflow Orchestration and State Transitions}

As illustrated in Figure~\ref{fig:architecture_flow}, our Agentic Lybic system operates through a self-consistent state-driven workflow that orchestrates seamless coordination between its four core components. The workflow begins when a user query enters the system, triggering the Central Controller to initialize the global state and enter the main execution pipeline.

\textbf{Initialization and Planning Phase}: Upon receiving a user query, the Controller sets the task state to "created" and transitions to the REPLAN situation ($R_p$). The Manager performs objective alignment to understand the user's intent, then conducts dynamic task planning to decompose the goal into executable subtasks. Using DAG generation, the Manager creates a structured plan where subtasks are organized with explicit dependencies. The Rule Engine monitors this process, ensuring planning attempts remain within configured limits (default: 10 attempts).

\textbf{Action Generation and Execution Cycle}: Once subtasks are available, the Controller transitions to GET\_ACTION ($G_a$). The Worker subsystem receives role assignments based on subtask characteristics—Operator for GUI interactions, Technician for system operations, or Analyst for decision support. Each Worker generates specific actions appropriate to their domain expertise. When Workers generate actual executable actions (such as Click, TypeText, or system commands), the Controller transitions to EXECUTE\_ACTION ($E_a$), where the Executor component coordinates hardware-level operation execution. However, when Workers return decision signals (CANNOT\_EXECUTE, SUPPLEMENT, STALE, or DONE), the Controller transitions to appropriate states based on the specific decision: CANNOT\_EXECUTE triggers REPLAN, SUPPLEMENT activates SUPPLEMENT state, STALE leads to QUALITY\_CHECK, and DONE proceeds to QUALITY\_CHECK for verification.

\textbf{Continuous Quality Monitoring}: After each execution step, the system enters QUALITY\_CHECK ($Q_c$) through multiple trigger mechanisms. The Evaluator employs three distinct triggers: periodic assessment every 5 steps, stagnation detection when identical actions repeat more than 3 times, and verification when Workers report task completion. The gate decision framework processes visual state comparisons and progress analysis to determine one of five outcomes: gate\_done (continue to next subtask), gate\_fail (trigger re-planning), gate\_continue (maintain current strategy), gate\_supplement (request additional information), or gate\_error (handle evaluation errors and trigger appropriate recovery mechanisms).

\textbf{Adaptive Coordination and Recovery}: When gate\_supplement is triggered, the Controller transitions to SUPPLEMENT ($S_p$), where the Manager queries external knowledge sources or requests clarification. Failed quality checks return the system to REPLAN, enabling the Manager to perform light, medium, or heavy adjustments based on failure severity. This adaptive mechanism allows the system to modify parameters, restructure dependencies, or completely re-decompose tasks as needed.

\textbf{Completion and Verification}: When all subtasks reach completion status, the Controller enters FINAL\_CHECK ($F_c$). The Evaluator performs comprehensive verification against the original objectives, ensuring all task requirements have been satisfied. Only upon successful final verification does the system transition to the terminal DONE state, marking task fulfillment.

The workflow implements robust error handling through state recovery mechanisms. Invalid states trigger transitions to INIT for system reset, while execution errors redirect to appropriate recovery paths. This design ensures the system maintains operational stability even when individual components encounter failures, providing graceful degradation rather than complete system breakdown. The Rule Engine continuously monitors system health through configurable thresholds: maximum state switches (default: 100), task runtime limits, and execution step boundaries. These safeguards prevent infinite loops and ensure resource-bounded operation while maintaining execution flexibility for complex tasks.

\section{Results}
\begin{table}[!h]
\caption{Comparison of the state-of-the-art methods on the OSWorld~\citep{xie2024osworldbenchmarkingmultimodalagents} benchmark. We show the approach type in the second column. A specialized model means that the model is trained specifically for computer use. We report the success rate (\%) as the evaluation metric in the third column. We only include the verified results from \url{https://os-world.github.io/}.}

\resizebox{\textwidth}{!}{
\setlength{\tabcolsep}{46pt}
    \centering
    \begin{tabular}{lcc}
        \toprule
        \textbf{Agent Model} & \textbf{Approach Type} & \textbf{Success Rate}\\
        \midrule
        \multicolumn{3}{l}{\textit{50 steps}} \\
        \midrule
        o3 \citep{openAI_o3_o4_mini} & General Model & 17.17 \\
        opencua-a3b \citep{wang2025opencua} & Specialized model & 19.93 \\
        opencua-qwen2-7b \citep{wang2025opencua} & Specialized model & 20.60 \\
        UI-TARS-72B-DPO \citep{qin2025ui} & Specialized Model & 25.80 \\
        Jedi-7B w/ gpt-4o \citep{xie2025scalingcomputerusegroundinguser}  & Agentic Framework & 26.92 \\
        UI-TARS-1.5-7B  \citep{qin2025ui} & Specialized Model & 27.30 ± 2.1  \\
        opencua-7b \citep{wang2025opencua} & Specialized model & 28.20±0.5 \\
        TianXi-Action-7B \citep{tang2025seaselfevolutionagentstepwise} & Specialized model & 29.80±0.6 \\
        OpenAI CUA 4o \citep{openAI_o3_o4_mini} & Specialized Model & 31.30 \\
        opencua-32b \citep{wang2025opencua} & Specialized Model & 34.10 ± 0.7 \\
        claude-3-7-sonnet-20250219 \citep{claude37} & General Model & 35.80 \\
        claude-4-sonnet-20250514 \citep{anthropic2025claude4} & General Model & 43.90 \\
        Agent S2 w/ Gemini-2.5-Pro \citep{agashe2025agents2compositionalgeneralistspecialist} & Agentic Framework & 45.76 \\
        autoglm-os-9b \citep{lai2025computerrl} & Specialized model & 47.26 \\
        GTA-1-7B w/ o3 \citep{yang2025gta1guitesttimescaling}  & Agentic Framework & 48.59 \\
        Jedi-7B w/ o3 \citep{xie2025scalingcomputerusegroundinguser} & Agentic Framework & 50.65  \\
        Agent S2.5 w/ o3 \citep{agashe2025agents2compositionalgeneralistspecialist} & Agentic Framework & 54.21 \\
        CoACT-1 w/ o3 \citep{song2025coact1computerusingagentscoding} & Agentic Framework & 56.39 \\
        \midrule
        \ours w/ o3 \& UI-TARS & Agentic Framework & \textbf{57.07} \\
        \midrule
    \end{tabular}
}

\label{tab:osworld_comparison}
\end{table}

\subsection{Experimental Setup}

\noindent\textbf{Benchmark and Dataset.} We evaluate Agentic Lybic on the OSWorld benchmark~\citep{xie2024osworldbenchmarkingmultimodalagents}, a scalable real-computer testbed that exposes a Linux OS environment to agents through both pixel streams and shell interfaces. OSWorld comprises 361 tasks (excluding Google Drive tasks) spanning common productivity tools, IDEs, browsers, file managers, and multi-application workflows, providing comprehensive coverage of vision-language grounding and long-horizon planning challenges in heterogeneous GUI environments. Each task includes: (i) a deterministic VM snapshot capturing the initial desktop state, (ii) natural-language goals mirroring end-user requests (e.g., "resize the image to $512\times512$ and export as PNG"), and (iii) rule-based evaluators built from 134 atomic execution-based components. Tasks range from atomic operations to complex cross-application pipelines, offering a realistic spectrum of automation complexity.

\noindent\textbf{Implementation Details.} We implement Agentic Lybic with careful selection of backbone models for each component. For all core reasoning components (Controller, Manager, and the three Worker roles - Technician, Operator, and Analyst), we utilize OpenAI o3~\citep{openAI_o3_o4_mini} to leverage its advanced reasoning capabilities for complex task orchestration and decision-making. For visual grounding and GUI action generation within the Operator worker, we employ UI-TARS~\citep{qin2025ui}, a specialized vision-language model specifically fine-tuned for computer use tasks and GUI element recognition. The Evaluator component also uses o3 for comprehensive quality assessment and gate decision-making. Our system operates with carefully tuned limits: quality checks triggered every 5 execution steps, stagnation detection after 3 consecutive identical actions, and re-planning triggered when single subtask execution exceeds 15 actions. Our implementation is publicly available at \url{https://github.com/xlang-ai/OSWorld/tree/main/mm_agents/maestro} to facilitate reproducibility and future research. The OSWorld official held the evaluation process and reported the verified result Success Rate.

\noindent\textbf{Evaluation Protocol.} We employ the rule-based evaluator provided by OSWorld, which expresses each task as Boolean expressions built from the 134 atomic evaluators. Task completion requires satisfying complex logical conditions (e.g., (file exported AND MD5 matches) AND (email sent == True)), ensuring comprehensive verification of task objectives rather than superficial completion signals.

\begin{table}[htbp]
\centering
\renewcommand{\arraystretch}{1.5}
\caption{\label{tab:main-res} Per-task performance comparison on OSWorld tasks with 50 steps budget. The number beside each subtask is the total number of that subtask. It is obvious that \ours outperforms the previous SOTA methods in almost all subtasks and achieves the best overall average performance. Subtasks with the most performance gain are Chrome, Impress, GIMP, and OS, demonstrating the effectiveness of our framework.}
\resizebox{\textwidth}{!}{%
\begin{tabular}{lccccccccccc}
\hline
\textbf{Methods (50 steps)}          & \textbf{Chrome (46)}    & \textbf{Calc (47)}      & \textbf{Impress (47)}   & \textbf{Writer (23)}    & \textbf{GIMP (26)}      & \textbf{VSCode (23)}   & \textbf{Multi Apps (93)} & \textbf{Thunderbird (15)} & \textbf{OS (24)}        &\textbf{ VLC (17)}       & \textbf{Avg.}           
\\ \hline
autoglm-os & 36.96          & 58.70          & 27.57          & 52.17          & 57.69          & 69.57          & 33.43             & 80.00            & 66.67          & 64.29          & 47.26         
\\ \hline
GTA-1         & 34.78          & 40.43          & 44.60          & 60.74 & 73.08          & \textbf{82.61}          & 37.05            & \textbf{86.67}            & 58.33          & 35.47          & 48.59  
\\ \hline
Jedi-7B  & 57.69          & 40.43          & 44.66          & 65.22          & 80.77          & 56.52          & 34.97             & 80.00            & 54.17          & 58.06          & 50.65          
\\ \hline
Agent S2.5         & 52.09          & 55.32          & 55.30          & 47.83 & 76.92          &73.91          &39.53            &73.33           & 75.00          & 42.00          & 54.21
\\ \hline
CoACT-1       & 45.57 & \textbf{68.09} & 46.72 & \textbf{73.91}          & 61.54 & 78.26 & \textbf{42.37}   & 66.67   & 70.83 & \textbf{66.06} & 56.39
\\ \hline
\ours       & \textbf{60.78} & 51.06 & \textbf{59.48} & 69.56          & \textbf{84.62} & 73.91 & 35.56   & 73.33   & \textbf{79.17} & 63.75 & \textbf{57.07} 
\\ \hline
\end{tabular}%
}
\end{table}

\subsection{Main Results}

\autoref{tab:osworld_comparison} presents comprehensive performance comparisons on OSWorld, demonstrating that Agentic Lybic establishes new state-of-the-art results across multiple evaluation settings. Our system achieves a remarkable success rate of \textbf{57.07\%} at 50 steps, substantially outperforming all existing methods including CoAct-1 (56.39\%), Agent S2.5 w/ o3 (54.21\%), and Jedi-7B w/ o3 (50.65\%). This represents a meaningful advancement in desktop automation capabilities, particularly considering the challenging nature of OSWorld tasks.

\autoref{tab:main-res} provides detailed breakdowns across application categories, revealing where our orchestration approach provides the most significant advantages. Agentic Lybic demonstrates strong performance across most application categories, with particularly notable results in Chrome browser tasks (60.78\% vs. CoAct-1's 45.57\%), LibreOffice Impress presentations (59.48\% vs. CoAct-1's 46.72\%), GIMP image editing (84.62\% vs. CoAct-1's 61.54\%), and OS-level operations (79.17\% vs. CoAct-1's 70.83\%).

\subsection{Efficiency Analysis}

Beyond success rates, Agentic Lybic demonstrates remarkable efficiency improvements through intelligent orchestration. Our system reduces the average number of steps required for task completion while maintaining higher success rates, indicating more effective task execution strategies. The tiered reasoning approach enables the system to select optimal execution modalities for each subtask, avoiding inefficient GUI manipulation sequences when programmatic approaches are more suitable.

\begin{figure}[!t]
  \includegraphics[width=\textwidth]{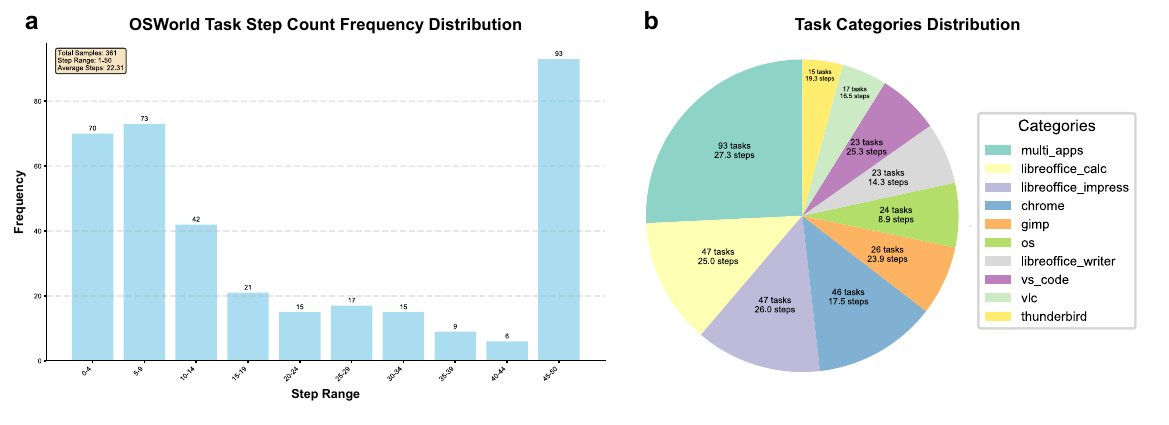}
  \caption{OSWorld benchmark task complexity analysis. (a) Step count frequency distribution showing task completion requirements across the dataset. The distribution reveals the diverse complexity range in OSWorld tasks, from simple atomic operations to complex multi-step workflows. (b) Task category distribution across the 361 OSWorld tasks, showing multi-application workflows comprise the largest portion (93 tasks, 25.2\% with average 27.3 steps), followed by LibreOffice applications (Calc: 47 tasks with 28.0 steps, Impress: 47 tasks with 28.0 steps, Writer: 23 tasks with 25.3 steps), Chrome browser tasks (46 tasks, 17.5 steps), and other specialized applications including GIMP (26 tasks, 20.9 steps), VS Code (23 tasks, 14.3 steps), OS operations (24 tasks, 8.8 steps), VLC (17 tasks, 10.7 steps), and Thunderbird (15 tasks, 16.7 steps).}
  \label{fig:step}
\end{figure}

The OSWorld benchmark presents diverse complexity challenges that highlight the effectiveness of our orchestration approach (Figure~\ref{fig:step}). The task complexity analysis reveals a heterogeneous distribution with 361 tasks ranging from simple atomic operations (0-4 steps, 70 tasks) to complex multi-step workflows (45-50 steps, 93 tasks), with an average of 22.31 steps per task. The task category distribution (Figure~\ref{fig:step}b) shows that multi-application workflows comprise the largest portion (93 tasks, 25.2\% with average 27.3 steps), followed by LibreOffice applications, Chrome browser tasks, and specialized tools. This complexity distribution demonstrates why sophisticated orchestration mechanisms are essential—simple GUI-only approaches may suffice for atomic operations, but complex workflows requiring 25+ steps (representing 38.8\% of all tasks) benefit significantly from our multi-modal coordination capabilities.

The quality gate system contributes significantly to efficiency by enabling early detection of execution issues and proactive course correction, preventing costly error propagation that typically requires extensive recovery procedures in other systems. Our periodic checks (every 5 steps), stagnation detection (3 consecutive identical actions), and success verification mechanisms ensure optimal resource utilization while maintaining robustness.

\subsection{Error Analysis and Robustness}

Analysis of failure cases reveals that Agentic Lybic maintains robustness across diverse error conditions while highlighting both system limitations and evaluation challenges.

\textbf{Evaluation Standard Limitations.} A significant portion of observed failures stem from overly rigid evaluation criteria rather than actual system deficiencies. For instance, in a calculation task requiring multiplication of total work hours by hourly rate, our agent correctly computed the result but formatted it to two decimal places, while the gold standard required exactly four decimal places for acceptance. Similarly, in a presentation-to-video conversion task, our agent successfully exported slides to PNG format and encoded them into MP4 using ffmpeg, yet failed evaluation because the gold standard incorrectly marked this task as impossible to complete, as shown in Figure~\ref{fig:fail}. These cases highlight the need for more flexible evaluation frameworks that assess functional correctness rather than rigid formatting requirements.

\textbf{System Robustness Indicators.} Despite these challenges, Agentic Lybic demonstrates superior error handling compared to baseline methods. The system's multi-modal execution strategy proves particularly effective in complex scenarios requiring seamless coordination across multiple applications. Figure~\ref{fig:success} illustrates a representative successful case where the agent handles a sophisticated multi-step workflow: extracting an AWS invoice PDF from a local email in the "Bills" folder, moving it to the receipts folder while following existing file naming patterns, and updating a tally book record. This example showcases the agent's capability to seamlessly coordinate across multiple applications (email client, file manager, spreadsheet) while maintaining context awareness for naming conventions and data entry patterns.

The incremental clarification policy effectively resolves visual ambiguities in GUI-dense environments, preventing accumulation of grounding errors that typically plague pure vision-based approaches. Stagnation detection triggers (activated after 3 consecutive identical actions) successfully prevent infinite loops, while periodic quality checks (every 5 steps) enable early intervention before critical failures occur. The tiered architecture enables graceful degradation when individual components encounter limitations. When the Operator struggles with visual grounding ambiguity, the system automatically transitions to Technician-based programmatic approaches where available. The Manager's adaptive re-planning capabilities (light, medium, and heavy adjustment levels) provide multiple recovery strategies based on failure severity, ensuring robust performance across diverse failure conditions.

Our analysis demonstrates that while Agentic Lybic faces challenges inherent to complex desktop automation, the sophisticated coordination mechanisms and continuous quality control provide substantial improvements in both success rates and error recovery compared to existing approaches. The multi-modal execution strategy, combined with robust quality gates and adaptive re-planning, enables the system to handle diverse failure conditions while maintaining operational effectiveness across complex multi-application workflows.

\section{Conclusion}

In this work, we introduced Agentic Lybic, a novel multi-agent execution system that addresses fundamental limitations in desktop automation through tiered reasoning architecture and sophisticated orchestration mechanisms. Our system advances beyond existing approaches by implementing a comprehensive four-tier framework comprising a Controller for state management, a Manager for intelligent task decomposition, specialized Workers for different execution modalities, and an Evaluator for continuous quality assessment.

The key innovation of our approach lies in the dynamic orchestration mechanism that seamlessly coordinates between GUI manipulation, system-level operations, and analytical decision-making based on real-time task assessment. Unlike previous systems that rely on static task delegation, Agentic Lybic implements continuous feedback loops through comprehensive quality gates, enabling adaptive re-planning and robust error recovery throughout task execution. This principled approach to multi-agent coordination, combined with specialized worker roles and incremental clarification policies, provides a more robust foundation for handling the inherent complexity of real-world desktop automation.

Our experimental evaluation on the challenging OSWorld benchmark demonstrates the effectiveness of this approach, achieving a new state-of-the-art success rate of 57.07\% in 50 steps—a substantial improvement over existing methods including CoAct-1 (56.39\%) and Agent S2.5 (54.21\%). Beyond raw performance gains, our system exhibits superior reliability across diverse task categories. These results validate our hypothesis that principled orchestration with continuous quality control provides significant advantages over both pure GUI agents and simpler hybrid approaches.

Our approach has certain inherent constraints: real-time visual understanding for tasks like video editing or gaming with continuous visual changes, and scenarios requiring human verification such as CAPTCHAs or secure authentication processes. Additionally, highly specialized software domains may require deeper contextual knowledge than current models provide.

Our work opens several promising directions for future research. The tiered reasoning framework provides a systematic foundation for incorporating additional specialized components, such as domain-specific workers for complex applications like video editing software or development environments. The quality gate system could be extended with more sophisticated intervention strategies, potentially including predictive error detection and proactive resource allocation. Furthermore, the orchestration mechanisms could be adapted to handle collaborative multi-user scenarios or distributed computing environments.

As vision-language models continue advancing, our orchestration framework provides a flexible foundation for integrating new capabilities into more robust automation systems. The success of our approach suggests that the future of desktop automation lies in developing sophisticated coordination mechanisms that can orchestrate multiple specialized components in principled and adaptive ways, opening possibilities for truly autonomous computing assistants that handle diverse human-computer interaction tasks with reliability and efficiency.




\bibliographystyle{ACM-Reference-Format}
\bibliography{main}

\appendix

\setcounter{figure}{0}
\setcounter{table}{0}
\renewcommand{\thefigure}{A.\arabic{figure}}
\renewcommand{\thetable}{A.\arabic{table}}

\section{System Trigger Codes}

This section provides a comprehensive overview of the trigger codes used in the \textsc{Agentic Lybic} system for state transitions and component coordination. These trigger codes enable precise tracking and debugging of system behavior across different execution phases.

\subsection{Trigger Code Categories}

The Controller employs a comprehensive trigger code system that enables precise coordination between components and robust error handling. Each component in the system communicates with the Controller through specific trigger codes that indicate the current execution status and determine the next appropriate controller situation. The trigger system is organized into ten primary categories:

\textbf{Rule Validation} triggers implement the foundation of our automated quality control system, providing continuous oversight that prevents system degradation and ensures optimal resource utilization. The rule\_quality\_check\_steps trigger enforces periodic assessment every 5 execution steps, creating a systematic evaluation rhythm that enables early detection of potential issues before they propagate into critical failures. When rule\_quality\_check\_repeated\_actions fires after detecting more than 3 identical consecutive actions, it indicates execution stagnation and immediately transitions the system to QUALITY\_CHECK state, preventing infinite loops and resource waste on ineffective strategies. The rule\_replan\_long\_execution trigger serves as an efficiency safeguard, activating when single subtask execution exceeds 15 actions to initiate strategic reconsideration through the REPLAN state, ensuring the system doesn't persist with suboptimal approaches when more effective alternatives may exist.

\begin{table}[!htbp]
\centering
\caption{Complete Trigger Code Reference for \textsc{Agentic Lybic} System}
\label{tab:trigger_codes}
\resizebox{\textwidth}{!}{%
\begin{tabular}{|l|l|p{6cm}|l|}
\hline
\textbf{Category} & \textbf{Trigger Code} & \textbf{Description} & \textbf{Target State} \\
\hline
\multirow{3}{*}{\textbf{Rule Validation}} 
& \texttt{rule\_quality\_check\_steps} & Periodic quality check every 5 steps & QUALITY\_CHECK \\
& \texttt{rule\_quality\_check\_repeated\_actions} & Triggered when identical actions repeated >3 times & QUALITY\_CHECK \\
& \texttt{rule\_replan\_long\_execution} & Single subtask execution exceeds 15 actions & PLAN \\
\hline
\multirow{5}{*}{\textbf{Task Status Rules}} 
& \texttt{rule\_max\_state\_switches\_reached} & Maximum state switches exceeded & DONE \\
& \texttt{rule\_plan\_number\_exceeded} & Planning attempts exceed threshold & DONE \\
& \texttt{rule\_state\_switch\_count\_exceeded} & State switch count limit reached & DONE \\
& \texttt{rule\_task\_completed} & Task successfully completed & DONE \\
& \texttt{rule\_task\_runtime\_exceeded} & Task runtime limit exceeded & DONE \\
\hline
\multirow{3}{*}{\textbf{INIT State}} 
& \texttt{subtask\_ready} & First subtask available for execution & GET\_ACTION \\
& \texttt{no\_subtasks} & No subtasks available, need planning & PLAN \\
& \texttt{init\_error} & Error during initialization & PLAN \\
\hline
\multirow{8}{*}{\textbf{GET\_ACTION State}} 
& \texttt{no\_current\_subtask\_id} & Missing current subtask identifier & INIT \\
& \texttt{subtask\_not\_found} & Referenced subtask not found & INIT \\
& \texttt{worker\_success} & Worker completed subtask successfully & QUALITY\_CHECK \\
& \texttt{work\_cannot\_execute} & Worker cannot execute current subtask & PLAN \\
& \texttt{worker\_stale\_progress} & Worker progress stagnated & QUALITY\_CHECK \\
& \texttt{worker\_supplement} & Worker requires additional information & SUPPLEMENT \\
& \texttt{worker\_generate\_action} & Worker generated new action & EXECUTE\_ACTION \\
& \texttt{no\_worker\_decision} & No decision from worker & PLAN \\
& \texttt{get\_action\_error} & Error during action generation & PLAN \\
\hline
\multirow{3}{*}{\textbf{EXECUTE\_ACTION}} 
& \texttt{execution\_error} & Error during action execution & GET\_ACTION \\
& \texttt{command\_completed} & Command executed successfully & GET\_ACTION \\
& \texttt{no\_command} & No command available for execution & GET\_ACTION \\
\hline
\multirow{5}{*}{\textbf{QUALITY\_CHECK}} 
& \texttt{all\_subtasks\_completed} & All subtasks finished & FINAL\_CHECK \\
& \texttt{quality\_check\_passed} & Quality assessment successful & GET\_ACTION \\
& \texttt{quality\_check\_failed} & Quality assessment failed & PLAN \\
& \texttt{quality\_check\_supplement} & Additional info needed & SUPPLEMENT \\
& \texttt{quality\_check\_execute\_action} & Additional execution required & EXECUTE\_ACTION \\
& \texttt{quality\_check\_error} & Error during quality check & PLAN \\
\hline
\multirow{2}{*}{\textbf{PLAN State}} 
& \texttt{subtask\_ready\_after\_plan} & New subtasks ready after planning & GET\_ACTION \\
& \texttt{plan\_error} & Error during planning phase & INIT \\
\hline
\multirow{2}{*}{\textbf{SUPPLEMENT}} 
& \texttt{supplement\_completed} & Information supplement finished & PLAN \\
& \texttt{supplement\_error} & Error during supplementation & PLAN \\
\hline
\multirow{4}{*}{\textbf{FINAL\_CHECK}} 
& \texttt{final\_check\_error} & Error during final verification & DONE \\
& \texttt{final\_check\_pending} & Additional subtasks discovered & GET\_ACTION \\
& \texttt{final\_check\_passed} & Final verification successful & DONE \\
& \texttt{final\_check\_failed} & Final verification failed & PLAN \\
& \texttt{task\_impossible} & Task determined impossible & DONE \\
\hline
\multirow{2}{*}{\textbf{Error Recovery}} 
& \texttt{unknown\_state} & Unrecognized system state & INIT \\
& \texttt{error\_recovery} & General error recovery & INIT \\
\hline
\end{tabular}%
}
\end{table}

\textbf{Task Status Rules} provide comprehensive system-wide oversight and resource management through monitoring mechanisms that enforce operational boundaries and detect completion conditions. These triggers collectively ensure that the system operates within defined computational limits while maintaining flexibility for complex task execution. The rule\_max\_state\_switches\_reached and rule\_state\_switch\_count\_exceeded triggers prevent excessive state transitions that could indicate system thrashing or infinite loops, terminating execution when predefined limits are exceeded. The rule\_plan\_number\_exceeded trigger stops planning attempts that surpass configured thresholds (typically 10 attempts), preventing the system from consuming excessive resources on intractable planning problems. Task completion is managed through rule\_task\_completed, which signals successful fulfillment and transitions to the terminal DONE state, while rule\_task\_runtime\_exceeded enforces temporal boundaries to ensure bounded execution regardless of task complexity.

\textbf{INIT State} triggers manage system initialization and task startup procedures, establishing the foundation for successful execution by ensuring proper system configuration and task availability. The subtask\_ready trigger indicates that the Manager has successfully created initial subtasks and the system can proceed to action generation through GET\_ACTION state. When no subtasks are available, the no\_subtasks trigger redirects the system to REPLAN state, ensuring that execution cannot proceed without proper task decomposition. The init\_error trigger handles initialization failures by transitioning to REPLAN state, enabling recovery from startup problems through strategy reconsideration and system reconfiguration.

\textbf{GET\_ACTION State} triggers coordinate the critical interface between task planning and execution, managing communication with Worker components and handling the diverse outcomes of action generation attempts. The worker\_success trigger signals successful subtask completion by any Worker component, immediately transitioning to QUALITY\_CHECK state for verification and progress assessment. When Workers encounter execution difficulties, work\_cannot\_execute indicates their inability to handle the current subtask, triggering a transition to REPLAN state for strategic adjustment. The worker\_stale\_progress trigger detects execution stagnation and routes to QUALITY\_CHECK for intervention, while worker\_supplement requests additional information through SUPPLEMENT state transition. Successful action generation is confirmed by worker\_generate\_action, enabling progression to EXECUTE\_ACTION state. Error conditions are handled through no\_worker\_decision and get\_action\_error, both triggering REPLAN state for recovery and strategy reassessment. Navigation errors are managed by no\_current\_subtask\_id and subtask\_not\_found, which reset the system to INIT state for proper reinitialization.

\textbf{EXECUTE\_ACTION} triggers manage the execution phase where generated actions are translated into actual system operations, providing essential feedback about command execution success and failure conditions. The command\_completed trigger confirms successful action execution and returns control to GET\_ACTION state for continuation with the next action in the sequence. When execution encounters problems, execution\_error redirects back to GET\_ACTION state, enabling the system to attempt alternative actions or request new action generation from Worker components. The no\_command trigger handles cases where no executable command is available, returning to GET\_ACTION state to ensure continuous progress and prevent execution stalls.

\textbf{QUALITY\_CHECK} triggers implement comprehensive evaluation mechanisms that assess execution effectiveness and determine optimal continuation strategies based on current progress and system state. The all\_subtasks\_completed trigger recognizes task completion across all subtasks and transitions to FINAL\_CHECK state for comprehensive verification. Successful progress continuation is managed by quality\_check\_passed, which returns to GET\_ACTION state for next subtask execution. When quality assessment identifies problems, quality\_check\_failed triggers REPLAN state transition for strategic adjustment and recovery. Information gaps are addressed through quality\_check\_supplement, which activates SUPPLEMENT state for additional context gathering. Direct execution requirements are handled by quality\_check\_execute\_action, enabling immediate transition to EXECUTE\_ACTION state when additional operations are needed. Error conditions during quality assessment are managed by quality\_check\_error, triggering REPLAN state for system recovery and strategy reassessment.

\textbf{PLAN State} triggers manage the planning and re-planning processes, ensuring that task decomposition and strategic adjustment proceed effectively while handling planning failures gracefully. The subtask\_ready\_after\_plan trigger indicates successful planning completion with new subtasks available for execution, enabling transition to GET\_ACTION state to begin or resume task execution. Planning failures are handled by plan\_error, which triggers INIT state transition for system reinitialization and recovery, ensuring that planning problems don't propagate through the system and cause cascading failures.

\textbf{SUPPLEMENT} triggers coordinate information gathering and external resource integration, enabling the system to address knowledge gaps and ambiguous situations through additional context acquisition. The supplement\_completed trigger signals successful information gathering and returns the system to REPLAN state, where the newly acquired information can be integrated into updated task strategies and decomposition. Supplementation failures are managed by supplement\_error, which also transitions to REPLAN state, ensuring that information gathering problems are addressed through strategic reconsideration rather than system stalls.

\textbf{FINAL\_CHECK} triggers implement comprehensive task verification and completion assessment, providing the final validation layer before task termination and handling discovery of additional requirements. The final\_check\_passed trigger confirms successful task completion according to all specified criteria and transitions to DONE state for system termination. When final verification reveals additional work requirements, final\_check\_pending redirects to GET\_ACTION state for continued execution. Completion assessment failures are managed by final\_check\_failed, triggering REPLAN state for final strategy adjustment. The task\_impossible trigger provides an explicit mechanism for recognizing intractable tasks, transitioning directly to DONE state to prevent infinite execution attempts. System errors during final verification are handled by final\_check\_error, also terminating in DONE state to ensure clean system shutdown.

\textbf{Error Recovery} triggers provide robust fault handling mechanisms that ensure system stability and graceful degradation under unexpected conditions or unrecognized states. The unknown\_state trigger handles situations where the system encounters unrecognized or invalid states, providing a safe fallback by transitioning to INIT state for clean reinitialization. The general error\_recovery trigger serves as a comprehensive safety mechanism for unexpected system errors, also redirecting to INIT state to ensure that unforeseen problems don't cause system crashes or undefined behavior, instead enabling graceful recovery through controlled reinitialization.

This trigger-driven architecture ensures that Agentic Lybic maintains precise control over execution flow while providing the flexibility necessary for handling diverse desktop automation scenarios. The comprehensive trigger code system enables both fine-grained component coordination and robust system-wide oversight, contributing significantly to the system's superior performance on complex long-horizon tasks.

\section{Case Study Analysis}
\begin{figure}[!b]
  \includegraphics[width=\textwidth]{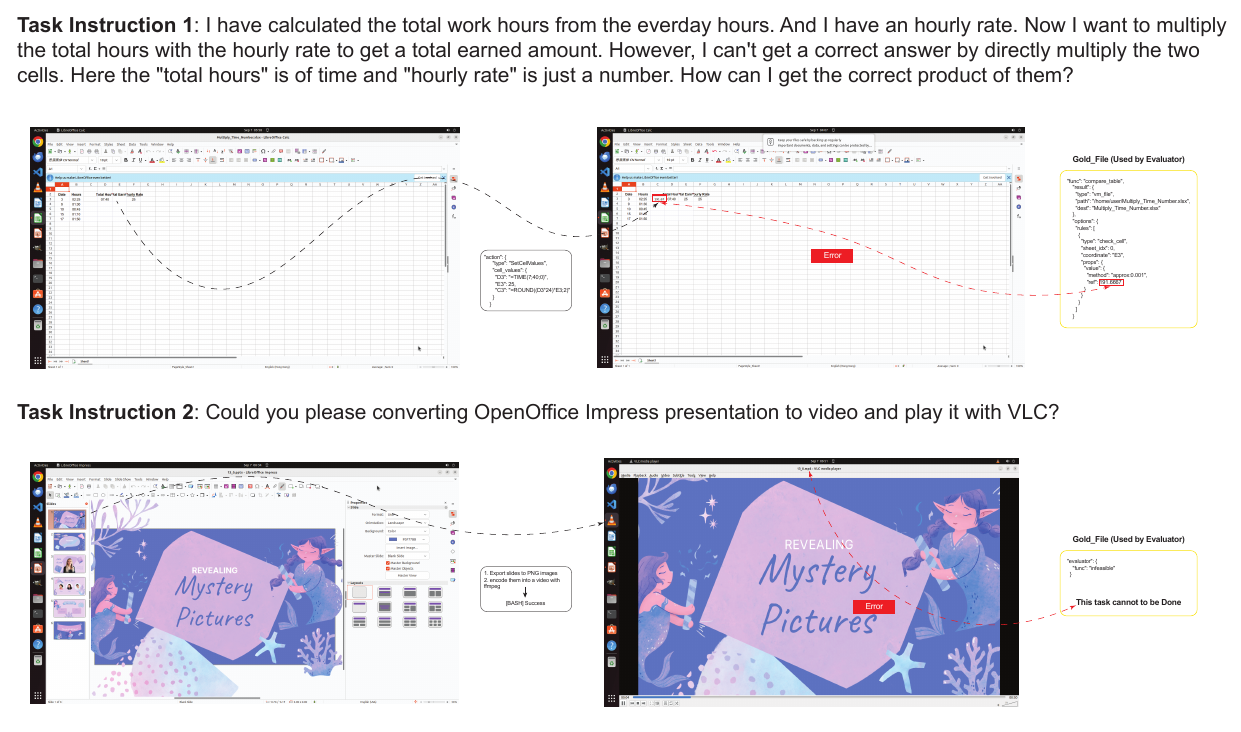}
  \caption{Failure case analysis demonstrating evaluation standard limitations. Task 1 shows a calculation task where the agent correctly computed the product of total hours and hourly rate but formatted the result to two decimal places, while the evaluation gold standard required four decimal places for correctness. Task 2 demonstrates a presentation-to-video conversion task where the agent successfully exported slides to PNG images and encoded them into an MP4 video using ffmpeg, but failed evaluation because the gold standard marked this task as impossible to complete. These cases highlight how rigid evaluation criteria can misclassify successful agent execution as failures.}
  \label{fig:fail}
\end{figure}

This section provides detailed analysis of representative success and failure cases to illustrate system behavior and evaluation challenges. Figure~\ref{fig:fail} demonstrates instances where the system functionally succeeded but was marked as failed due to overly rigid evaluation standards, including a calculation task where correct computation was rejected for decimal formatting and a video conversion task that was completed successfully but deemed "impossible" by the evaluator. These cases reveal important limitations in current benchmark evaluation methodologies that may underestimate actual system capabilities. 

\begin{figure}[!t]
  \includegraphics[width=\textwidth]{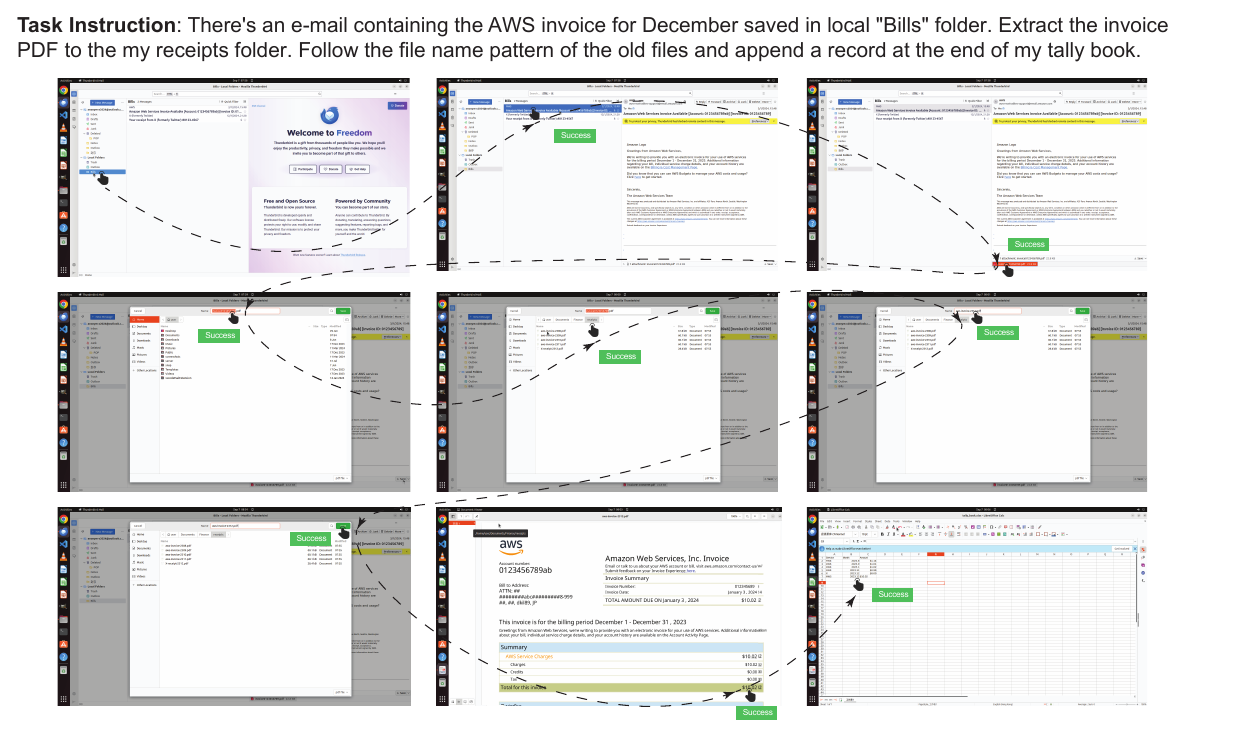}
  \caption{Successful case demonstrating multi-modal task execution. The task involves extracting an AWS invoice PDF from a local email in the "Bills" folder, moving it to the receipts folder following existing file naming patterns, and updating a tally book record. This example showcases the agent's capability to seamlessly coordinate across multiple applications (email client, file manager, spreadsheet) while maintaining context awareness for naming conventions and data entry patterns.}
  \label{fig:success}
\end{figure}

In contrast, Figure~\ref{fig:success} showcases a complex multi-modal workflow where Agentic Lybic demonstrates sophisticated coordination capabilities by seamlessly orchestrating operations across email client, file manager, and spreadsheet applications while maintaining contextual awareness for file naming conventions and data entry patterns. This successful case exemplifies the system's ability to handle intricate cross-application workflows that require sustained coordination and context preservation across multiple execution phases.

\end{document}